\documentclass[runningheads]{llncs}
\usepackage[T1]{fontenc}
\usepackage{multirow} 
\usepackage{booktabs} 
\usepackage{xcolor}
\usepackage{colortbl}
\usepackage{graphicx,verbatim}
\usepackage{amsmath} 
\usepackage{amssymb}
\usepackage{subcaption}
\usepackage[colorlinks=true,citecolor=green]{hyperref}
\usepackage[misc]{ifsym}
\newcommand{\para}[1]{\vspace{.05in}\noindent\textbf{#1}}

\begin{document}
\title{Recover Semantics First, Generate Better: Improved Latent Modeling for 3D MRI Reconstruction and Cross-Contrast Synthesis}
\titlerunning{Semantics-First Latent Modeling for 3D MRI}

%

\author{
Yonghao Chen\inst{1,2}\textsuperscript{*} \and
Sicheng Yang\inst{1}\textsuperscript{*} \and
Rui Tang\inst{1} \and
Lei Zhu\inst{1}\textsuperscript{\Letter}
}

\institute{
The Hong Kong University of Science and Technology (Guangzhou)
\and
Xi'an Jiaotong University 
}

\maketitle

\begingroup
\renewcommand{\thefootnote}{\fnsymbol{footnote}}
\footnotetext[1]{Equal contribution.}
\endgroup

\begingroup
\renewcommand{\thefootnote}{\Letter}
\footnotetext{Lei Zhu (leizhu@hkust-gz.edu.cn) is the corresponding author.}
\endgroup

\begin{abstract}
Multi-contrast magnetic resonance imaging (MRI) provides complementary information for clinical diagnosis. However, acquiring all MRI sequences is often time-consuming and costly. Recent generative models perform cross-contrast synthesis to address this issue by inferring absent contrasts from the available ones. Nevertheless, synthesizing 3D MRI presents significant challenges. Due to the massive volume sizes, operating directly in the pixel space is computationally prohibitive; therefore, a common approach is to first compress the 3D volumes into a latent space and subsequently train generative models in that space. We observe that existing compression architectures face several critical issues: they under-preserve long-range anatomical coherence, discard clinically meaningful semantics, and rely on optimization objectives that lead to over-smoothed reconstructions. Ultimately, these shortcomings compromise the performance of subsequent generative models. In this work, we propose a semantics-first latent modeling framework for 3D MRI reconstruction and cross-contrast synthesis. Specifically, we introduce a Latent Harmonization Encoder (LHE) to capture global anatomical dependencies, ensuring coherent volumetric representations. To mitigate semantic degradation during latent compression, we further design a Semantic Recovery Block (SRB) that injects high-level priors from a self-supervised semantic teacher, enhancing contrast-aware separability in the latent space. Additionally, we propose an Anatomy-aware Frequency Loss (AFL) to adaptively preserve diagnostically relevant high-frequency structures. Extensive experiments on two public multi-contrast MRI datasets demonstrate consistent improvements in reconstruction fidelity and cross-contrast synthesis quality.
Our code is available at \url{https://github.com/script-Yang/RSF}.
\keywords{Latent representation learning \and 3D MRI synthesis}
\end{abstract}

\section{Introduction}
Multi-contrast MRI provides complementary information vital for diagnosis, but acquiring all sequences is resource-intensive, driving the need for cross-contrast synthesis~\cite{brown2014magnetic,kavur2021chaos,zuo2023haca3,arslan2025self,kui2025comprehensive}. While clinical applications increasingly demand \textbf{3D MRI synthesis} to preserve spatial continuity, massive volume sizes (e.g., $256 \times 256 \times 256$) make pixel-space generation computationally prohibitive~\cite{rombach2022high,khader2023denoising}. Consequently, current methods adopt a hierarchical approach: compressing 3D volumes into a latent space via VAE/VQ models~\cite{kingma2013auto,van2017neural}, followed by latent-space generation via GANs~\cite{goodfellow2020generative} or Diffusion~\cite{rombach2022high}. Despite its efficiency, applying this pipeline to 3D MRI presents three primary hurdles.

\textbf{(1) Long-range anatomical incoherence:} Anatomical structures in 3D MRI exhibit strict spatial continuity across the volume. However, conventional latent models often fail to capture these long-range dependencies, leading to structural distortions and disjointed anatomy in the generated volumes~\cite{taleb20203d,hatamizadeh2021swin}.
\textbf{(2) Semantic entanglement across contrasts:} Different MRI protocols encode distinct contrast-specific semantics. Existing compressors often entangle these nuances, causing a semantic degradation that impairs downstream generative modeling~\cite{dalmaz2022resvit}.
\textbf{(3) Over-smoothing from pixel-wise objectives:} Standard reconstruction losses inherently prioritize low-frequency accuracy, yielding overly smooth representations that obscure clinically vital fine-grained details, such as subtle boundaries or small lesions~\cite{isola2017image,jiang2021focal}.

To address these challenges, we propose a unified framework centered on the principle of \emph{recovering semantics first to enable better generation}. Our framework integrates three cooperative components: (1) a \textbf{Latent Harmonization Encoder (LHE)} that facilitates global information interaction to ensure anatomical coherence; (2) a \textbf{Semantic Recovery Block (SRB)} that leverages high-level self-supervised priors to enhance contrast-specific discriminability; and (3) an \textbf{Anatomy-aware Frequency Loss (AFL)} that adaptively enforces high-frequency fidelity in clinically relevant regions. Extensive experiments on two public datasets demonstrate that our approach achieves state-of-the-art performance in both 3D MRI reconstruction and multi-contrast synthesis.

\begin{figure*}[t]
    \centering
     \includegraphics[width=0.95\textwidth]{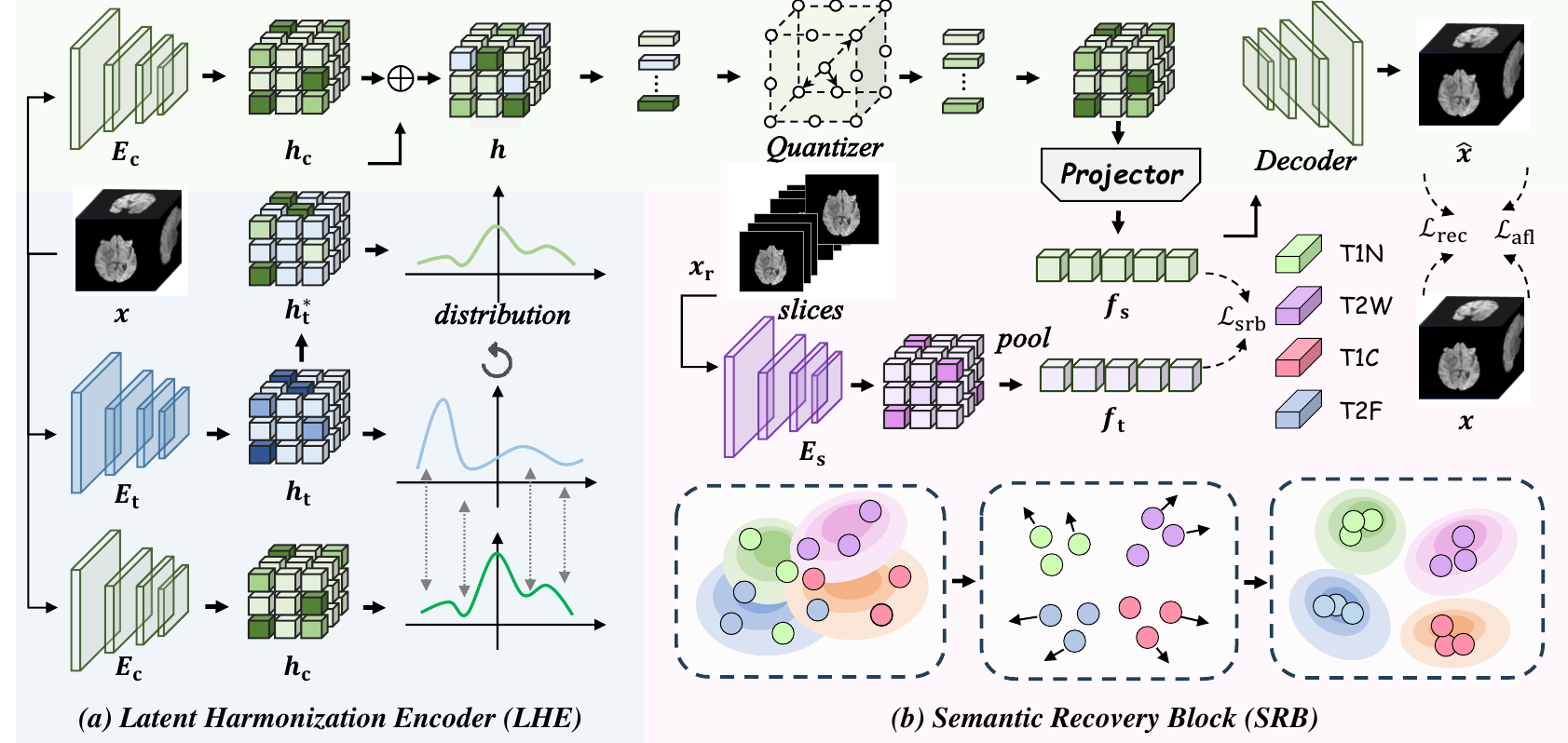}
    \caption{Overview of our proposed semantics-first latent modeling framework. The model compresses 3D MRI volumes into a structured latent space using: (a) a Latent Harmonization Encoder (LHE), which captures long-range anatomical dependencies via heterogeneous encoders $E_\text{c}$ and $E_\text{t}$; (b) a Semantic Recovery Block (SRB), which aligns latent features $f_\text{s}$ with high-level priors $f_\text{t}$ from a self-supervised teacher to restore lost semantic information and ensure contrast-aware separability; and (c) an Anatomy-aware Frequency Loss (AFL, detailed in Fig.~\ref{fig:afl}) to mitigate the over-smoothing caused by reconstruction losses.}
    \label{fig:framework_overview}
\end{figure*}

\section{Methodology}
\subsection{Overall Framework}
As shown in Figs.~\ref{fig:framework_overview} and~\ref{fig:afl}, our framework consists of three cooperative components: 
1) a Latent Harmonization Encoder (LHE), which captures long-range anatomical dependencies; 
2) a Semantic Recovery Block (SRB), which re-injects high-level semantic guidance into the latent representation; and 
3) an Anatomy-aware Frequency Loss (AFL), which enforces high-frequency consistency in clinically important regions.

\subsection{Latent Harmonization Encoder (LHE)}
In 3D MRI synthesis, purely convolutional encoders often fail to capture the long-range spatial coherence of anatomical structures, leading to inconsistencies~\cite{wang2023swinmm,hatamizadeh2021swin}. To address this, we introduce a context-aware harmonization stage \emph{before} latent discretization. Given an input volume $x$, we extract local convolutional features $h_{\text{c}} = E_{\text{c}}(x)$. To capture global dependencies, we employ a parallel, slice-wise Vision Transformer (ViT)~\cite{dosovitskiy2020image} to extract context features $h_{\text{t}} = E_{\text{t}}(x)$. 

To mitigate statistical mismatches between the two distinct pathways, we apply channel-wise feature alignment prior to fusion:
\begin{equation}
h_{\text{t}}^{*} = \frac{h_{\text{t}} - \mu_{\text{t}}}{\sigma_{\text{t}} + \epsilon} \cdot \sigma_{\text{c}} + \mu_{\text{c}},
\end{equation}
where $\mu$ and $\sigma$ denote channel-wise mean and standard deviation. The aligned ViT features are then integrated via residual fusion: $h = h_{\text{c}} + h_{\text{t}}^{*}$. 

Finally, the fused representation $h$ is discretized via finite scalar quantization (FSQ)~\cite{mentzer2023finite,agarwal2025cosmos}. Quantizing these context-aware features ensures that the discrete latent codes retain long-range anatomical relationships, preventing the structural fragmentation often seen in standard compression.

\subsection{Semantic Recovery Block (SRB)}
\label{sec:SRB}
\begin{figure}[t]
    \centering
    \includegraphics[width=\textwidth]{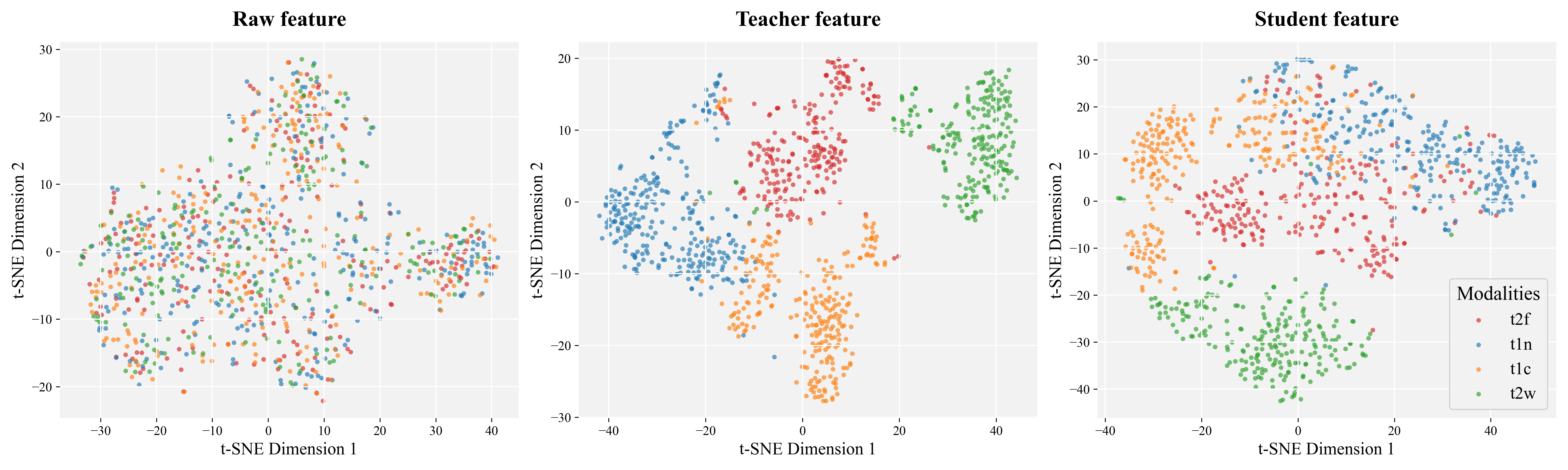}
    \caption{T-SNE visualization of latent representations. (a) Raw FSQ latents exhibit severe entanglement, hindering cross-contrast learning. (b) DINO teacher features $f_\text{t}$ show distinct contrast-wise clustering. (c) Our Semantic Recovery Block (SRB) aligns student embeddings $f_\text{s}$ with the teacher, restoring semantic separability and enhancing the generative model's cross-contrast mapping.}
    \label{fig:srb_tsne}
\end{figure}

Standard VAE/VQ-based compression prioritizes low-level fidelity over high-level anatomical semantics~\cite{van2017neural,yang2026vaevq,yang2026vq}. In multi-contrast MRI, this causes latent entanglement (Fig.~\ref{fig:srb_tsne}(a)), impeding cross-contrast correspondence learning. To mitigate this, our Semantic Recovery Block (SRB) re-injects semantic priors into the latent space via a self-supervised teacher~\cite{caron2021emerging,yu2024representation,wu2026representation}.

As shown in Fig.~\ref{fig:srb_tsne}(b), modern self-supervised models naturally extract contrast-specific features that exhibit clear clustering. Leveraging this, we feed a randomly sampled 2D slice $x_r$ from the input volume $x$ into a frozen DINO-pretrained ViT-B/16~\cite{caron2021emerging}. The teacher semantic embedding $f_t$ is obtained via global average pooling of normalized final-layer patch tokens:
\begin{equation}
f_\text{t} = \text{Pool}(\text{E}_{\text{s}}(x_\text{r})).
\end{equation}
Simultaneously, the quantized latent $z$ is mapped into the same semantic space through a learnable MLP projector $P(\cdot)$ to produce the student vector $f_\text{s}$:
\begin{equation}
f_\text{s} = P(z).
\end{equation}
Alignment is enforced via the semantic latent alignment loss:
\begin{equation}
\mathcal{L}_{\text{srb}} = \left\| f_{\text{s}} - f_{\text{t}} \right\|_2^2.
\end{equation}

By optimizing $\mathcal{L}_{\text{srb}}$, the latent embeddings achieve superior contrast-wise separability (Fig.~\ref{fig:srb_tsne}(c)), facilitating the learning of more robust cross-contrast relations for downstream synthesis.

\subsection{Anatomy-aware Frequency Loss (AFL)}
\begin{figure}[t]
    \centering
    \begin{minipage}{0.67\textwidth}
        \centering
        \includegraphics[width=\linewidth]{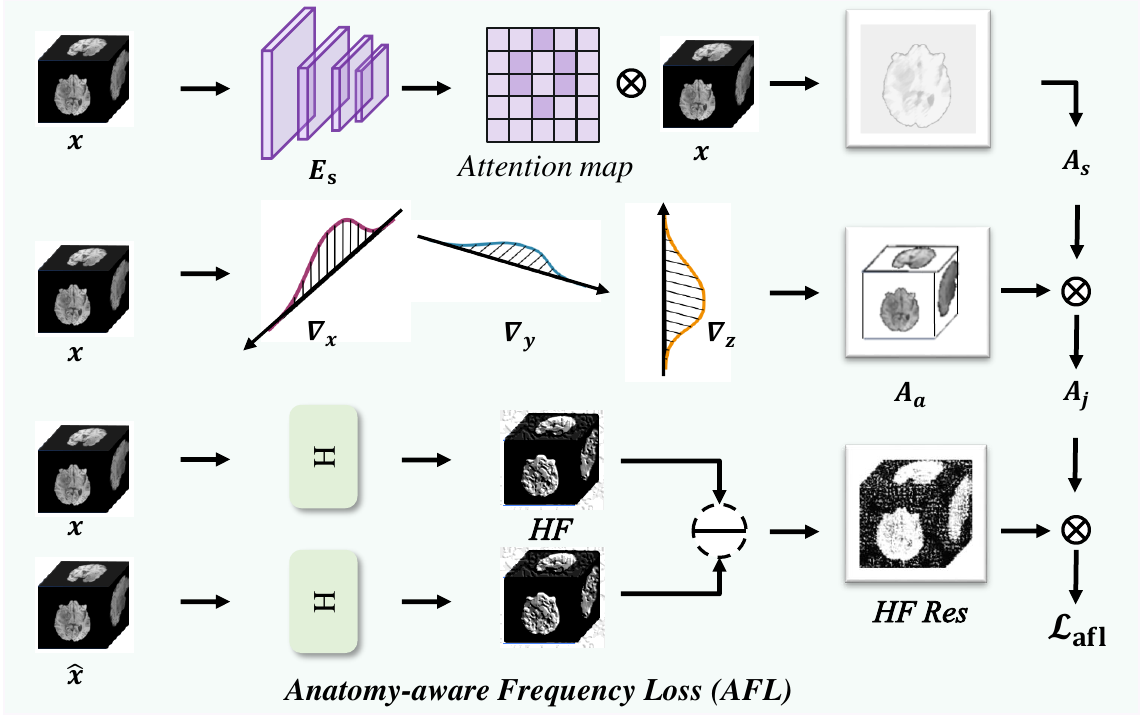}
        \caption{Illustration of the proposed Anatomy-aware Frequency Loss (AFL) mechanism.}
        \label{fig:afl}
    \end{minipage}%
    \hfill
    \begin{minipage}{0.29\textwidth}
        \centering
        \includegraphics[width=\linewidth]{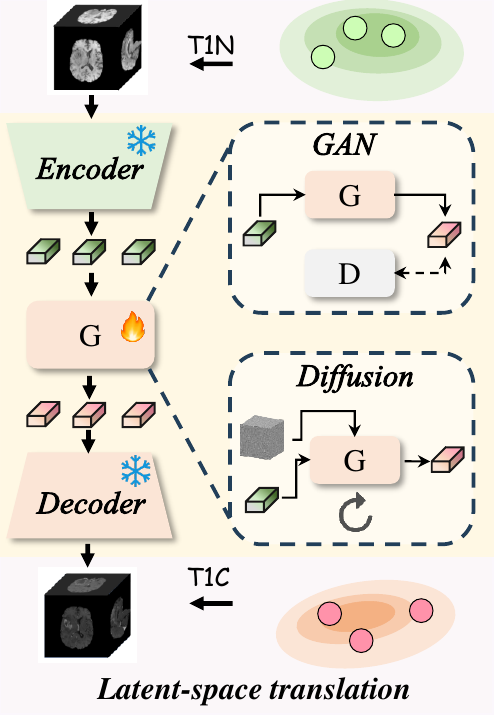}
        \caption{Latent cross-contrast synthesis.}
        \label{fig:lst}
    \end{minipage}
\end{figure}
Standard volumetric reconstruction objectives typically employ Mean Squared Error (MSE) loss, which prioritizes low-frequency components and often results in over-smoothed outputs~\cite{nie2016estimating,ledig2017photo}. To preserve high-frequency details essential for clinical diagnosis, we propose the Anatomy-aware Frequency Loss (AFL). As illustrated in Fig.~\ref{fig:afl}, AFL adaptively enforces high-frequency consistency using a joint anatomy-semantic attention mechanism.

To emphasize diagnostically meaningful regions, we construct a joint attention weight $A_{j} = A_{a} \odot A_{s}$. This is derived from two parallel streams. First, the anatomical attention map $A_{a}$ captures morphological discontinuities using volumetric spatial gradients:
\begin{equation}
A_{a} = \lvert \nabla_x x \rvert + \lvert \nabla_y x \rvert + \lvert \nabla_z x \rvert,
\end{equation}
where $\nabla_x$, $\nabla_y$, and $\nabla_z$ denote the first-order partial derivatives along the three spatial axes. Essentially, $A_{a}$ acts as a local edge detector to explicitly highlight fine structural boundaries and subtle textures. Second, the semantic attention map $A_{s}$ incorporates high-level structural relevance by aggregating and upsampling the final-layer self-attention responses from the pre-trained teacher network described in Section~\ref{sec:SRB}. This provides a global semantic filter, ensuring the high-frequency preservation prioritizes clinically significant tissues while suppressing irrelevant background artifacts.

Concurrently, we isolate the high-frequency residual $\mathcal{H}(x)$ of the input volume by subtracting its low-pass filtered representation:
\begin{equation}
\mathcal{H}(x) = x - \mathcal{S}_{v}(x),
\end{equation}
where $\mathcal{S}_{v}(\cdot)$ denotes a local volumetric smoothing operator. Finally, the AFL objective computes the $L_1$ distance between the attention-weighted high-frequency components of the reconstruction $\hat{x}$ and the target $x$:
\begin{equation}
\mathcal{L}_{\text{afl}} = \left\| A_{j} \odot \mathcal{H}(\hat{x}) - A_{j} \odot \mathcal{H}(x) \right\|_1.
\end{equation}
By restricting the high-frequency penalty specifically to anatomically and semantically significant regions, AFL effectively mitigates over-smoothing while maintaining robustness.

\subsection{Training Objective}
\label{sec:train_obj}

The final training objective is
\begin{equation}
\mathcal{L}_{\text{total}} =
\mathcal{L}_{\text{rec}} +
\mathcal{L}_{\text{srb}} +
\mathcal{L}_{\text{afl}}.
\end{equation}

Here, $\mathcal{L}_{\text{rec}}$ is the baseline reconstruction loss consistent with 3D FSQ~\cite{mentzer2023finite,agarwal2025cosmos}, 
while $\mathcal{L}_{\text{srb}}$ and $\mathcal{L}_{\text{afl}}$ are designed to enforce semantic consistency and maintain diagnostically relevant high-frequency structures, respectively.

\section{Experiments}
\begin{table*}[t]
\small
\centering
\caption{Quantitative comparison of reconstruction performance.}
\label{tab:recon_results}
\begin{tabular}{l|ccc|ccc}
\hline
& \multicolumn{3}{c|}{BraTS Avg} 
& \multicolumn{3}{c}{IXI Avg} \\
Method 
& PSNR $\uparrow$ & SSIM $\uparrow$ & LPIPS $\downarrow$ 
& PSNR $\uparrow$ & SSIM $\uparrow$ & LPIPS $\downarrow$ \\
\hline

VQVAE~\cite{van2017neural} & 27.43 & 0.8615 & 0.1342 & 28.61 & 0.8804 & 0.1128 \\
VQGAN~\cite{esser2021taming} & 29.84 & 0.8953 & 0.0816 & 31.06 & 0.9127 & 0.0705 \\
FSQ~\cite{agarwal2025cosmos}   & 32.39 & 0.9244 & \textbf{0.0557} & 33.18 & 0.9305 & 0.0514 \\
\rowcolor{gray!20} Ours  & \textbf{32.80} & \textbf{0.9281} & 0.0578 & \textbf{33.65} & \textbf{0.9377} & \textbf{0.0450} \\

\hline
\end{tabular}
\end{table*}
\begin{table*}[t]
\centering
\caption{Quantitative comparison on BraTS for two translation tasks.}
\label{tab:trans_results}
\small
\begin{tabular}{l|ccc|ccc}
\hline
& \multicolumn{3}{c|}{T2 $\rightarrow$ FLAIR} 
& \multicolumn{3}{c}{T1 $\rightarrow$ T1C} \\
Method 
& PSNR $\uparrow$ & SSIM $\uparrow$ & LPIPS $\downarrow$ 
& PSNR $\uparrow$ & SSIM $\uparrow$ & LPIPS $\downarrow$ \\
\hline

3D CycleGAN~\cite{zhu2017unpaired} (Pixel) 
& 26.45 & 0.8751 & 0.1423 
& 25.12 & 0.8224 & 0.1985 \\
Latent CycleGAN~\cite{zhu2017unpaired} (Baseline)
& 26.61 & 0.8669 & 0.1342 
& 26.93 & 0.8937 & 0.1272 \\

Latent Diffusion~\cite{rombach2022high} (Baseline)
& 25.17 & 0.7721 & 0.2276 
& 25.44 & 0.7706 & 0.2478 \\

\rowcolor{gray!20} Latent CycleGAN (Ours)
& \textbf{26.65} & \textbf{0.8769} & \textbf{0.1229} 
& \textbf{30.41} & \textbf{0.9252} & \textbf{0.0851} \\

\rowcolor{gray!20} Latent Diffusion (Ours)
& \textbf{27.95} & \textbf{0.7831} & \textbf{0.1623} 
& \textbf{26.55} & \textbf{0.7789} & \textbf{0.1874} \\

\hline
\end{tabular}
\end{table*}
\subsection{Datasets}
We evaluate our framework on two public multi-contrast brain MRI datasets: \textbf{BraTS}~\cite{baid2021rsna,labella2024analysis}, comprising T1-weighted (T1), post-contrast T1 (T1C), native T2-weighted (T2), and T2 fluid attenuated inversion recovery (FLAIR) sequences; and the \textbf{IXI} dataset~\cite{heckemann2006automatic}, comprising healthy T1, T2, and PD sequences. For both datasets, all 3D volumes are resampled to a uniform spatial resolution. We use their respective official data splits for training, validation, and testing.

\subsection{Implementation Details}
Our framework is trained in two stages, both conducted on 8 RTX4090 GPUs. For latent space construction, we employ 3D FSQ~\cite{mentzer2023finite,agarwal2025cosmos} as the baseline, with $E_\text{c}$ and the decoder adopting its structure. To evaluate the effectiveness of our proposed framework, we conduct comprehensive comparisons on two primary tasks: 3D MRI reconstruction and cross-contrast synthesis. We use PSNR, SSIM, and LPIPS to quantitatively assess both structural fidelity and perceptual quality.

\para{Stage 1: 3D MRI Reconstruction.}
In the first stage, we optimize the model with the reconstruction objective in Sec.~\ref{sec:train_obj} to learn compact latent representations. We use the Adam optimizer with an initial learning rate of $1 \times 10^{-4}$ and a total batch size of 2, and train the model for 500 epochs. After convergence, the modules responsible for compression and reconstruction (i.e., the encoder/decoder and quantization components) are frozen. To evaluate reconstruction fidelity, we compare our structurally enhanced latent space against representative volumetric compression methods, including VQ-VAE~\cite{van2017neural}, VQGAN~\cite{esser2021taming}, and vanilla FSQ~\cite{agarwal2025cosmos}. 

\para{Stage 2: Cross-Contrast Synthesis in Latent Space.}
In the second stage, we perform latent-space translation for contrast-to-contrast synthesis (specifically, T1 $\rightarrow$ T1C and T2 $\rightarrow$ FLAIR), as illustrated in Fig.~\ref{fig:lst}. We use the Adam optimizer with an initial learning rate of $1 \times 10^{-4}$ and a batch size of 32, and train the translation model for 200 epochs. Additionally, we employ an $L_1$ loss as the translation objective. Importantly, our goal in this stage is not to propose a novel generative architecture, but rather to demonstrate that a well-structured, semantically separable latent space can significantly boost the performance of existing generative models. To this end, we select CycleGAN~\cite{zhu2017unpaired} and Latent Diffusion Models~\cite{rombach2022high} as our GAN and diffusion-based comparison methods.

\begin{figure}[t]
    \centering
    \begin{minipage}{0.48\textwidth}
        \centering
        \includegraphics[width=\linewidth]{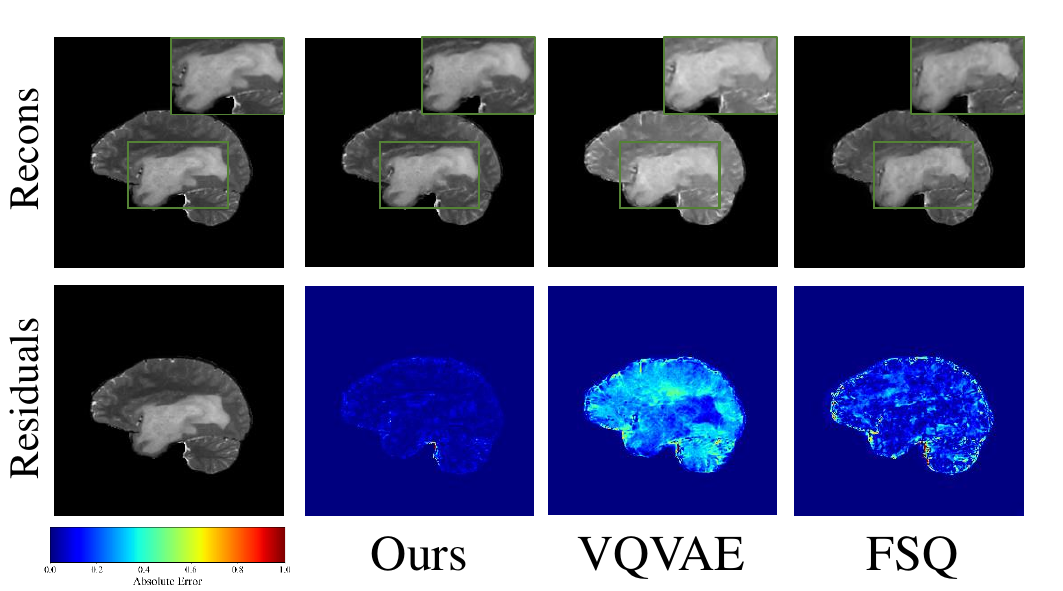}
        \caption{Visual comparison of 3D MRI reconstruction performance.}
        \label{fig:recon}
    \end{minipage}%
    \hfill
    \begin{minipage}{0.48\textwidth}
        \centering
        \includegraphics[width=\linewidth]{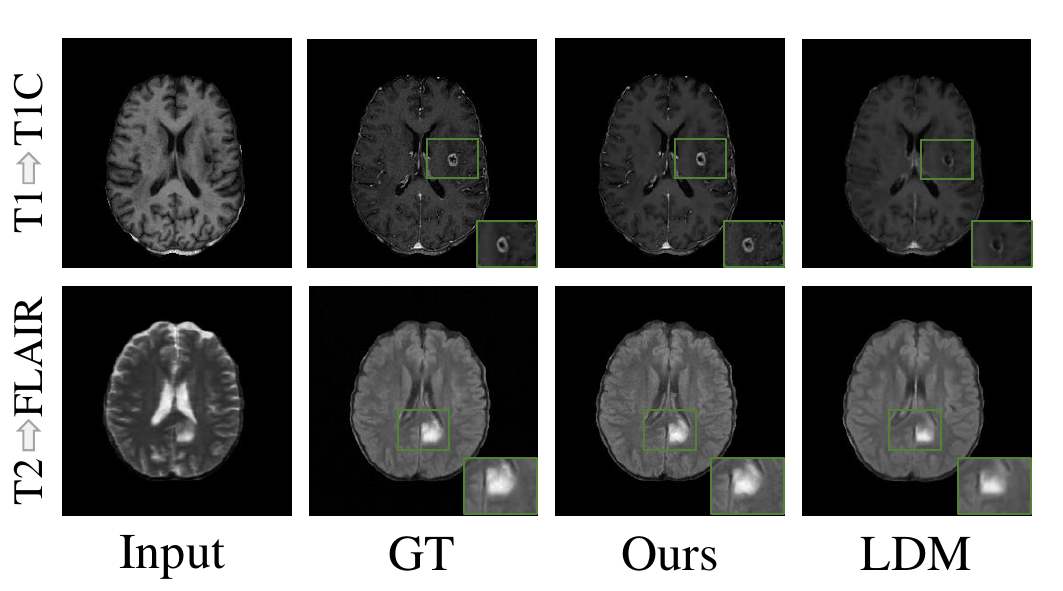}
        \caption{Visual comparison of cross-contrast synthesis results.}
        \label{fig:syn}
    \end{minipage}
\end{figure}
\subsection{Reconstruction and Translation Results}
Our method achieves state-of-the-art performance across both reconstruction and cross-contrast translation tasks. As summarized in Table~\ref{tab:recon_results}, it improves PSNR by 0.41 dB  over the strongest baseline on BraTS, alongside reaching 33.65 PSNR and the lowest LPIPS of 0.0450 on IXI. Furthermore, integrating this space into GAN and diffusion frameworks consistently outperforms existing baselines in Table~\ref{tab:trans_results}; notably, it yields a PSNR increase of 3.48 dB for Latent CycleGAN (T1$\rightarrow$T1C) and 2.78 dB for Latent Diffusion (T2$\rightarrow$FLAIR), proving that these gains stem from the semantically aligned latent representation. Qualitative results in Fig.~\ref{fig:recon} and Fig.~\ref{fig:syn} further illustrate our model's ability to recover fine-grained details and maintain structural consistency across contrasts.

\subsection{Ablation Studies}
\begin{table*}[t]
\centering
\small
\caption{Ablation study of the proposed modules.}
\label{tab:ablation_module}
\begin{tabular}{ccc|cccc|ccc}
\hline
LHE & SRB & AFL 
& \multicolumn{4}{c|}{PSNR $\uparrow$} 
& PSNR $\uparrow$
& SSIM $\uparrow$ 
& LPIPS $\downarrow$ 
 \\

& & 
& T1 & T2 & FLAIR & T1C 
& Avg & Avg & Avg \\
\hline

$\times$ & $\times$ & $\times$ 
& 31.60 & 30.38 & 32.88 & 34.69 
& 32.39 & 0.9234 & \textbf{0.0557} \\

$\checkmark$ & $\times$ & $\times$ 
& 31.98 & 31.05 & 32.86 & 34.71 
& 32.65 &  0.9256& 0.0558 \\

$\checkmark$ & $\checkmark$ & $\times$ 
& 31.78 & 30.91 & 32.73 & 34.45 
& 32.47 &  0.9253& 0.0563 \\

\rowcolor{gray!20} $\checkmark$ & $\checkmark$ & $\checkmark$ 
 & \textbf{32.30} & \textbf{31.20} & \textbf{33.01} & \textbf{34.69} 
& \textbf{32.80} &  \textbf{0.9281}& 0.0578 \\

\hline
\end{tabular}
\end{table*}

\begin{table*}[t]
\small
\centering
\caption{Ablation study on semantic teachers.}
\label{tab:ablation_encoder}
\begin{tabular}{l|cccc|ccc}
\hline
& \multicolumn{4}{c|}{PSNR $\uparrow$} &  PSNR $\uparrow$&  SSIM $\uparrow$&LPIPS $\downarrow$ \\
Encoder & T1 & T2 & FLAIR & T1C & Avg & Avg & Avg \\
\hline
ResNet-50
& 31.92 & 31.06 & 32.79 & 34.47 
& 32.56 & 0.9268 & 0.0594 \\

MAE (ViT-B)
& 31.75 & 31.00 & 32.76 & 34.33 
& 32.46 &  0.9261 & 0.0595\\

\rowcolor{gray!20} DINO (ViT-B/16)
& \textbf{32.30} & \textbf{31.20} & \textbf{33.01} & \textbf{34.69}
& \textbf{32.80} & \textbf{0.9281} & \textbf{0.0578} \\
\hline
\end{tabular}
\end{table*}
\para{Module Analysis}
As shown in Table \ref{tab:ablation_module}, our progressive ablation study on BraTS evaluates LHE, SRB, and AFL against a vanilla 3D FSQ baseline~\cite{agarwal2025cosmos}. Adding LHE improves average PSNR (32.39 to 32.65) by capturing long-range anatomical dependencies. Introducing SRB slightly trades pixel-level reconstruction
fidelity for improved semantic organization. Finally, integrating AFL achieves the peak performance of 32.80 dB PSNR and 0.9281 SSIM. Overall, these complementary modules combine to yield the strongest reconstruction results.

\para{Effect of Semantic Teachers.}
We adopt a DINO-pretrained ViT-B/16~\cite{caron2021emerging} as the semantic teacher due to its strong global representation capabilities~\cite{baharoon2023evaluating,song2024dino,yang2025segdino}. To validate this design choice, we compare it against pre-trained ResNet-50~\cite{he2016deep} and MAE (ViT-B) encoders~\cite{he2022masked} under identical settings on the BraTS dataset. As shown in Table~\ref{tab:ablation_encoder}, the DINO-based teacher achieves the best performance, obtaining the highest average PSNR (32.80) and SSIM (0.9281) alongside the lowest LPIPS (0.0578). These results confirm that DINO provides more robust and semantically aligned representations.
In future work, we will explore whether VLMs/LLMs~\cite{yang2026lcm} can provide stronger semantic supervision.

\section{Conclusion}
In this paper, we propose a semantics-first latent modeling framework for 3D MRI reconstruction and cross-contrast synthesis. Our method improves the structural coherence and semantic organization of compressed latents before generative modeling. To this end, we introduce a Latent Harmonization Encoder (LHE) to capture long-range anatomical dependencies, a Semantic Recovery Block (SRB) to enhance contrast-specific separability, and an Anatomy-aware Frequency Loss (AFL) to preserve diagnostically important details. This design reduces structural inconsistency and semantic entanglement in conventional VAE/VQ-based pipelines. Experiments on two public datasets demonstrate consistent improvements over volumetric compression and latent translation baselines, with ablations confirming the contribution of each component.

\begin{credits}
\subsubsection{\ackname}
This work was supported by the National Natural Science Foundation of China (Project No.82572383) and the Guangdong Science and Technology Department (2024ZDZX2004).
\subsubsection{\discintname}
The authors have no competing interests to declare that are relevant to the content of this article.
\end{credits}
%
%
%
%
%

\bibliographystyle{splncs04}
\end{document}